\documentclass[runningheads]{llncs}

\usepackage{times}
\usepackage{epsfig}
\usepackage{graphicx}
\usepackage{amsmath}
\usepackage{amssymb}
\usepackage{mathrsfs}
\usepackage{tabulary}
\usepackage{adjustbox}
\usepackage{caption}
\usepackage{graphicx}
\usepackage{caption}
\usepackage{subcaption}
\usepackage{floatrow}
\usepackage{xcolor}
\usepackage{wrapfig}

\newcolumntype{K}[1]{>{\centering\arraybackslash}p{#1}}
\newcommand\eg{{\emph{e.g.}}}
\newcommand\ie{{\emph{i.e.}}}
\newcommand\etal{{\emph{et al. }}}

\begin{document}
	\pagestyle{headings}
	\mainmatter

	\def\GCPR17SubNumber{24}

	\title{Motion Deblurring in the Wild}

	\titlerunning{Motion Deblurring in the Wild}
	\authorrunning{Noroozi \etal}
	
	\author{Mehdi Noroozi, Paramanand Chandramouli, Paolo Favaro}
	\institute{Institute for Informatics \\ University of Bern \\ \textbf{\texttt{\{noroozi, chandra, paolo.favaro\}@inf.unibe.ch} } }

	\maketitle

\begin{abstract}
We propose a deep learning approach to remove motion blur from a single image captured \emph{in the wild}, \ie, in an uncontrolled setting. Thus, we consider motion blur degradations that are due to both camera and object motion, and by occlusion and coming into view of objects. In this scenario, a model-based approach would require a very large set of parameters, whose fitting is a challenge on its own. Hence, we take a data-driven approach and design both a novel convolutional neural network architecture and a dataset for blurry images with ground truth. 
The network produces directly the sharp image as output and is built into three pyramid stages, which allow to remove blur gradually from a small amount, at the lowest scale, to the full amount, at the scale of the input image.
To obtain corresponding blurry and sharp image pairs, we use videos from a high frame-rate video camera. For each small video clip we select the central frame as the sharp image and use the frame average as the corresponding blurred image. Finally, to ensure that the averaging process is a sufficient approximation to real blurry images we estimate optical flow and select frames with pixel displacements smaller than a pixel. We demonstrate state of the art performance on datasets with both synthetic and real images.
\end{abstract}

\section{Introduction}
This work is concerned with the removal of blur in real images. We consider the challenging case where objects move in an arbitrary way with respect to the camera, and might be occluded and/or come into view. Due to the complexity of this task, prior work has looked at specific cases, where blur is the same everywhere (the shift-invariant case), see \eg, \cite{iccpsun,irani}, or follows given models \cite{Hyun2014,sun2015learning} and scenarios \cite{HuDepth,paramCVPR,Whyte2010}.
\begin{figure*}[t]
  \centering
  \begin{minipage}[b]{.245\textwidth}
        \includegraphics[width= 1 \textwidth,height=1.3\textwidth]{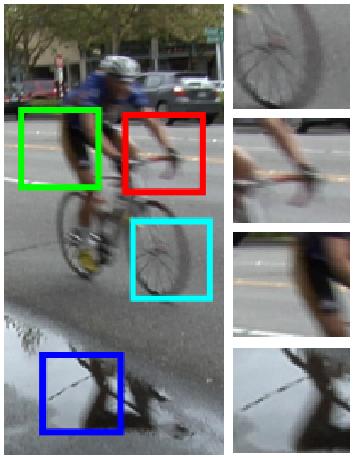}
   \end{minipage}
   \begin{minipage}[b]{.245\textwidth}
        \includegraphics[width= 1 \textwidth,height=1.3\textwidth]{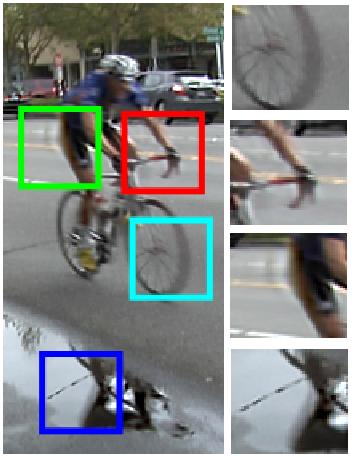}
   \end{minipage} 
   \begin{minipage}[b]{.245\textwidth}
        \includegraphics[width= 1 \textwidth,height=1.3\textwidth]{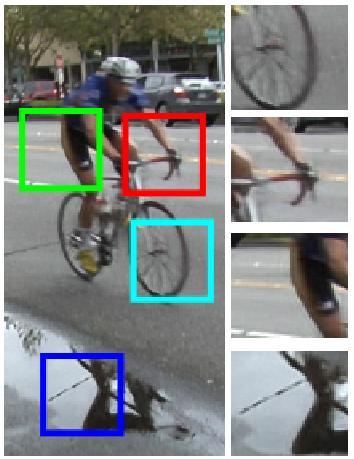}
   \end{minipage}   
   \begin{minipage}[b]{.245\textwidth}
        \includegraphics[width= 1 \textwidth,height=1.3\textwidth]{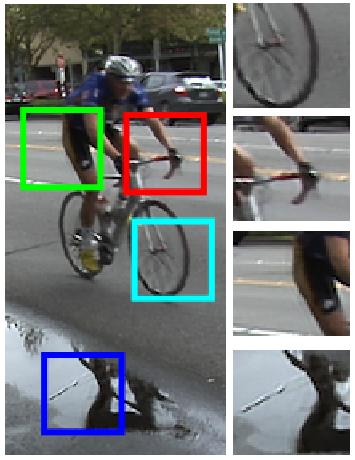}
   \end{minipage}\\
   (a)\hspace{.2\textwidth}(b)\hspace{.2\textwidth}(c)\hspace{.2\textwidth}(d)
  \caption{(a) Blurry video frame. (b) Result of \cite{sun2015learning} on the \emph{single} frame (a). (c) Result of the proposed method on the \emph{single} frame (a). (d) Result of the multi-frame method \cite{hyun2015generalized}. \label{fig:demo1}}
\end{figure*}
Other methods address the modeling complexity by exploiting multiple frames, as in, for example, \cite{hyun2015generalized}. 
Our objective, however, is to produce high-quality results as in \cite{hyun2015generalized} by using just a single frame (see Fig.~\ref{fig:demo1}). 
To achieve this goal we use a data-driven approach, where a convolutional neural network is trained on a large number of blurred-sharp image pairs. This approach entails addressing two main challenges: first, the design of a realistic dataset of blurred-sharp image pairs and second, the design of a suitable neural network that can learn from such dataset. We overcome the first challenge by using a commercial high frame-rate video camera (a GoPro Hero5 Black). Due to the high frame-rate, single frames in a video are sharp and motion between frames is small. Then, we use the central frame as the sharp image and the average of all the frames in a video clip as the corresponding blurry image. To avoid averaging frames with too much motion, which would correspond to unrealistic motion blurs, we compute the optical flow between subsequent frames and use a simple thresholding strategy to discard frames with large displacements (more than $1$ pixel). As we show in the Experiments section, a dataset built according to this procedure allows training a neural network and generalizes to images from other camera models and scenes.
To address the second challenge, we build a neural network that replicates (scale-space) pyramid schemes used in classical deblurring methods. 
The pyramid exploits two main ideas: one is that it is easy to remove a small amount of blur, and the second is that downsampling can be used to quickly reduce the blur amount in a blurry image (within some approximation). The combination of these two contributions leads to a method achieving state of the art performance on the single image space-varying motion blur case.

\subsection{Related work}

\noindent\textbf{Camera Motion.}
With the success of the variational Bayesian approach of Fergus et al.~\cite{Fergus2006}, a large number of blind deconvolution algorithms have been developed for motion deblurring~\cite{Babacan2012,Cho2009,Levin2011,iccpsun,Perrone2014,irani,xu2013unnatural,Xu2010}. Although blind deconvolution algorithms consider blur to be uniform across the image, some of the methods are able to handle small variations due to camera shake~\cite{Kohler2012}. Techniques based on blind deconvolution have been adapted to address blur variations due to camera rotations by defining the blur kernel on a higher dimensional space~\cite{Gupta2010,Hirsch2010,Whyte2010}. Another approach to handle camera shake induced space-varying blur is through region-wise blur kernel estimation~\cite{hirsch2011fast,ji2012two}. In 3D scenes, motion blur at a pixel is also related to its corresponding depth. To address this dependency, Hu~\etal and Xu and Jia \cite{HuDepth,Xu2012} first estimate a depth map and then solve for the motion blur and the sharp image. In~\cite{zheng2013forward}, motion blur due to forward or backward camera motion has been explicitly addressed. 
Notice that blur due to moving objects (see below) cannot be represented by the above camera motion models.

\noindent\textbf{Dynamic Scenes.} This category of blur is the most general one and includes motion blur due to camera or object motion. Some prior work \cite{Levin2006,couzinie2013learning} addresses this problem by assuming that the blurred image is composed of different regions within which blur is uniform. 
Techniques based on alpha matting have been applied to restore scenes with two layers~\cite{dai2009removing,tai2010coded}. Although these methods can handle moving objects, they require user interaction and cannot be used in general scenarios where blur varies due to camera motion and scene depth.
The scheme of Kim \etal~\cite{Hyun2013} incorporates alternating minimization to estimate blur kernels, latent image, and motion segments. Even with a general camera shake model for blurring, the algorithm fails in certain scenarios such as forward motion or depth variations~\cite{Hyun2014}. In~\cite{Hyun2014} Kim and Lee, propose a segmentation-free approach but assume a uniform motion model. The authors propose to simultaneously estimate motion flow and the latent image using a robust total variation (TV-L1) prior. Through a variational-Bayesian formulation, Schelten and Roth~\cite{schelten2014localized} recover both defocus as well as object motion blur kernels. Pan \etal~\cite{pansoft} propose an efficient algorithm to jointly estimate object segmentation and camera motion by incorporating soft segmentation, but require user input. 
\cite{chakrabarti2010analyzing,gast2016parametric,shi2014discriminative} address the problem of segmenting an image into different regions according to blur. 
Recent works that use multiple frames are able to handle space-varying blur quite well \cite{hyun2015generalized,WieSchLenHir16}. 
 
\noindent\textbf{Deep Learning Methods.}
The methods in~\cite{Schuler_2013_CVPR,NIPS2014_5485} address non-blind deconvolution wherein the sharp image is predicted using the blur estimated from other techniques. In~\cite{SchHirHarSch16}, Schuler \etal develop an end-to-end system that learns to perform blind deconvolution. Their system consists of modules to extract features, estimate the blur and to perform deblurring. 
However, the performance of this approach degrades for large blurs. The network of Chakrabarti~\cite{ayaneccv} learns the complex Fourier coefficients of a deconvolution filter for an input patch of the blurry image. 
Hradi{\v{s}} \etal~\cite{hradivs2015convolutional} predict clean and sharp images from text documents that are corrupted by motion blur, defocus and noise through a convolutional network without an explicit blur estimation. This approach has been extended to license plates in~\cite{license}. \cite{XiaWanHeiHir16} proposes to learn a multi-scale cascade of shrinkage fields model. 
This model however does not seem to generalize to natural images.
Sun \etal~\cite{sun2015learning} propose to address non-uniform motion blur represented in terms of motion vectors. 

Our approach is based on deep learning and on a single input image. However, we directly output the sharp image, rather than the blur, do not require user input and work directly on real natural images in the dynamic scene case. Moreover, none of the above deep learning methods builds a dataset from a high frame-rate video camera. Finally, our proposed scheme achieves state of the art performance in the dynamic scene case.

\section{Blurry Images in the Wild}
\label{sec:dataset}

One of the key ingredients in our method is to train our network with an, as much as possible, realistic dataset, so that it can generalize well on new data. As mentioned before, we use a high resolution high frame-rate video camera. We build blurred images by averaging a set of frames. Similar averaging of frames has been done in previous work to obtain data for evaluation~\cite{kim2016dynamic,agrawal2009optimal}, but not to build a training set. \cite{kim2016dynamic} used averaging to simulate blurry videos, and \cite{agrawal2009optimal} used averaging to synthesize blurry images, coded exposure images and motion invariant photographs. 

\begin{figure*}[t]
  \centering
  \begin{minipage}[b]{.48\textwidth}
        \includegraphics[width= 1 \textwidth, height=.5 \textwidth]{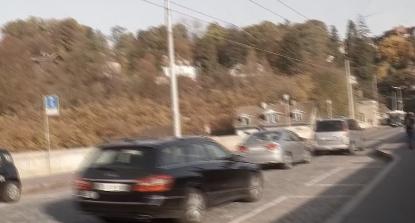}
   \end{minipage}
   \begin{minipage}[b]{.48\textwidth}
        \includegraphics[width= 1 \textwidth, height=.5 \textwidth]{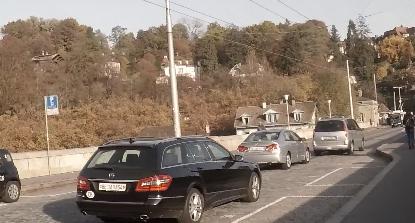}
   \end{minipage}\\
  \caption{A sample image pair from the WILD training set. Left: averaged image (the blurry image). Right: central frame (the sharp image).} 
  \label{fig:flow}
\end{figure*}

We use a handheld GoPro Hero5 Black camera, which captures $240$ frames per second with a resolution of $1280{\times}720$ pixels. Our videos have been all shot outdoors. Firstly, we downsample all the frames in the videos by a factor of $3$ in order to reduce the magnitude of relative motion across frames. Then, we select the number $N_e$ of averaged frames by randomly picking an odd number between $7$ and $23$. Out of the $N_e$ frames, the central frame is considered to be the sharp image. 
We assume that motion is smooth and, therefore, to avoid artifacts in the averaging process we consider only frames where optical flow is no more than $1$ pixel. We evaluate optical flow using the recent FlowNet algorithm \cite{DFIB15} and then apply a simple thresholding technique on the magnitude of the estimated flow. Fig.~\ref{fig:flow} shows an example of the sharp and blurred image pair in our training dataset. In this scene, we find both the camera and objects to be moving. 
We also evaluate when the optical flow estimate is reliable by computing the frame matching error ($L^2$ norm on the grayscale domain). 
We found that no frames were discarded in this processing stage (after the previous selection step).
We split our \emph{WILD} dataset into training and test sets.


\section{The Multiscale Convolutional Neural Network}
\begin{figure}[t]
\includegraphics[width=.8\linewidth,trim={2.5cm 1.8cm 2.1cm 6cm},clip=true]{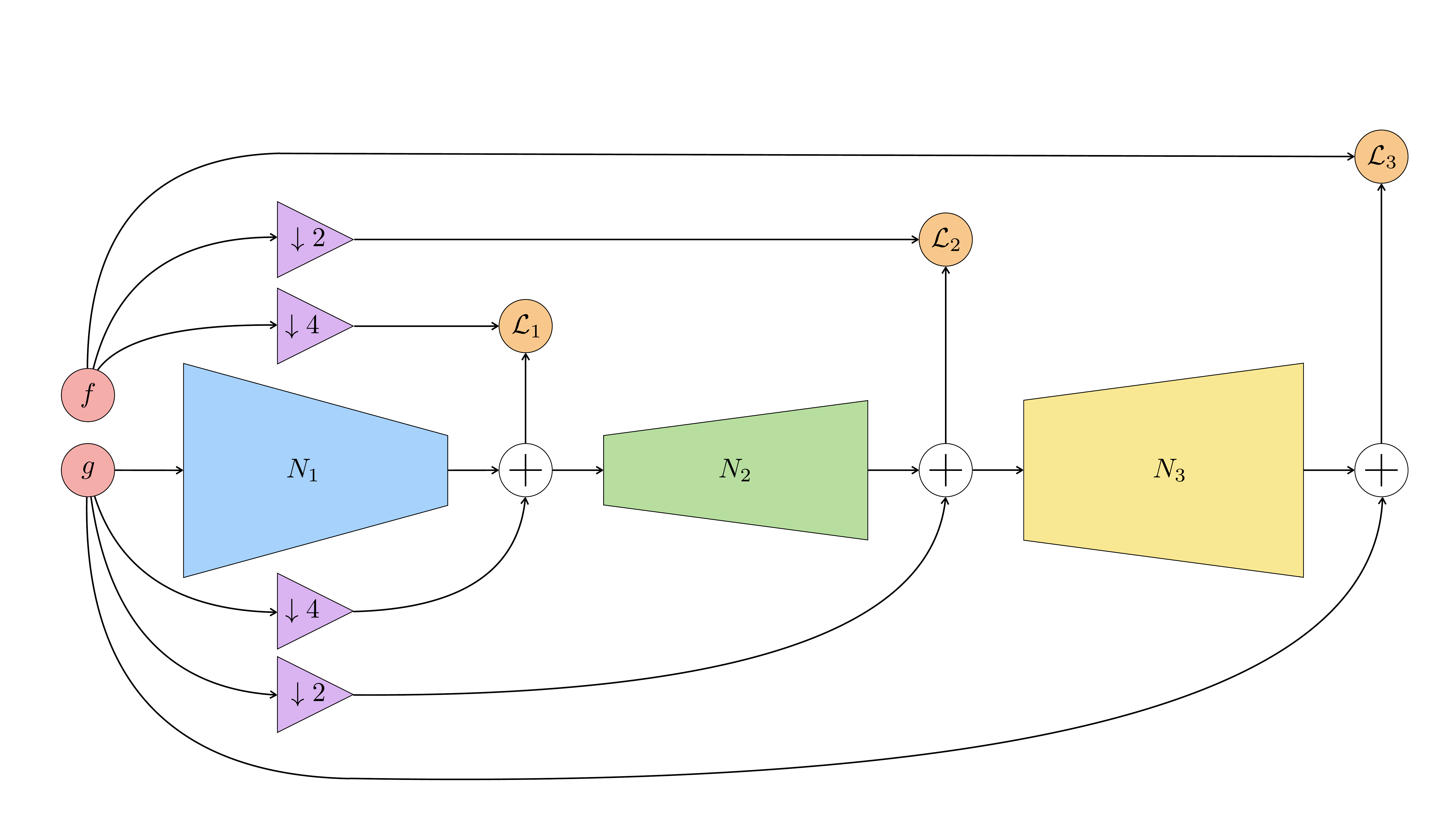} 
\caption{The DeblurNet architecture. The multiscale scheme allows the network to handle large blurs. Skip connections (bottom links) facilitate the generation of details. \label{fig:cnn}}
\end{figure}

In Fig.~\ref{fig:cnn} we show our proposed convolutional neural network (CNN) architecture. 
The network is designed in a pyramid or multi-scale fashion. Inspired by the multi-scale processing of blind deconvolution algorithms \cite{irani,SchHirHarSch16}, we introduce three subgraphs $N_1$, $N_2$, and $N_3$ in our network, where each subgraph includes several convolution/deconvolution 
(fractional stride convolution) layers. The task of each subgraph is to minimize the reconstruction error at a particular scale. 
There are two main differences with respect to conventional CNNs, which play a significant role in generating sharp images without artifacts. Firstly, the network includes a skip connection at the end of each subgraph. The idea behind this technique is to reduce the difficulty of the reconstruction task in the network by using the information already present in the blurry image. Each subgraph needs to only generate a \emph{residual image}, which is then added to the input blurry image (after downsampling, if needed). We observe experimentally that the skip connection technique helps the network in generating more texture details. 
Secondly, because the extent of blur decreases with downsampling \cite{irani}, the multi-scale formulation allows the network to deal with small amounts of blur in each subgraph. In particular, the task for the first subgraph $N_1$ is to generate a deblurred image residual at 1/4 of the original scale. The task for the subgraph $N_2$ is to use the output of $N_1$ added to the downsampled input and generate a sharp image at 1/2 of the original resolution. Finally, the task for the subgraph $N_3$ is to generate a sharp output at the original resolution by starting from the output of $N_2$ added to the input scaled by $1/2$. 
We call this architecture the \emph{DeblurNet} and give a detailed description in Tab.~\ref{tbl:cnn}.

\begin{table*}[t]
\begin{center}
\begin{adjustbox}{width=1\textwidth}
\begin{tabular}{ | l | c  c  c  c  c  c  c  | c  c  c  c  c | c  c  c  c  c | }
\hline
& \multicolumn{7}{c|}{$N_1$} &  \multicolumn{5}{c|}{$N_2$} & \multicolumn{5}{c|}{$N_3$}\\
\hline
Type  & conv & conv & conv & conv & conv  & conv  & conv &  conv & conv & conv & conv & deconv &  conv & conv & conv & conv & deconv\\ 
 OutCh  &  96 & 256	& 384 &384& 256&256 & 3 &  256 & 256 & 256 & 256 & 3 &  256 & 256 & 256 & 256 & 3 \\  
 Kernel  &  11& 7& 7& 7& 3&3&3&  5&5&5&5&5&  5&5&5&5&5\\
 Stride  &  $\downarrow$ 2& 1& 1& $\downarrow$ 2& 1&1& 1 & 1 & 1& 1& 1& $\uparrow$ 2 & 1 & 1& 1& 1& $\uparrow$ 2 \\
 \hline
\end{tabular}
\end{adjustbox}\vspace{-.3cm}
 \caption{The DeblurNet architecture. Batch normalization and ReLU layers inserted after every convolutional layer (except for the last layer of $N_1$) are not shown for simplicity. Downsampling ($\downarrow$) is achieved by using a stride greater than $1$ in convolutional layers. A stride greater than $1$ in deconvolutional ($\uparrow$) layers performs upsampling.} \label{tbl:cnn}
\end{center}
\end{table*}

\noindent\textbf{Training.} We minimize the reconstruction error of all the scales simultaneously. 
The loss function ${\cal L}= {\cal L}_1+{\cal L}_2+{\cal L}_3$ is defined through the following $3$ losses
\begin{equation}
\begin{split}
{\cal L}_1 &\textstyle = \sum_{(g,f)\in{\mathscr{D}}} \left|N_1(g) +  D_{\frac{1}{4}}(g) - D_{\frac{1}{4}}(f)\right|^2\\
{\cal L}_2 &\textstyle = \sum_{(g,f)\in{\mathscr{D}}} \left|N_2\left(N_1(g)+D_{\frac{1}{4}}(g)\right) + D_{\frac{1}{2}}(g) - D_{\frac{1}{2}}(f)\right|^2\\
{\cal L}_3 &\textstyle = \sum_{(g,f)\in{\mathscr{D}}} \left|N_3\left(N_2\left(N_1(g)+D_{\frac{1}{4}}(g)\right)+D_{\frac{1}{2}}(g)\right) + g - f\right|^2
\end{split}
\end{equation}
where $\mathscr{D}$ is the training set,  $g$ denotes a blurry image, $f$ denotes a sharp image,  $D_{\frac{1}{k}}(x)$ denotes the downsampling operation of the image $x$ by factor of $k$, and $N_i$ indicates the $i$-th subgraph in the DeblurNet, which reconstructs the image at the $i$-th scale. 

\noindent\textbf{Implementation Details.} We used Adam~\cite{kingma2014adam} for optimization with momentum parameters as $\beta_1= 0.9$, $\beta_2 = 0.999$, and an initial learning rate of $0.001$. We decrease the learning rate by $.75$ every $10^4$ iterations. We used 2 Titan X for training with a batch size of $10$. The network needs 5 days to converge using batch normalization~\cite{ioffe2015batch}. 


 

\section{Experiments}

We tested DeblurNet on three different types of data: a) the WILD test set (GoPro Hero5 Black), b) real blurry images (Canon EOS 5D Mark II), and c) data from prior work. 

\noindent\textbf{Synthetic vs pseudo-real training.}
To verify the impact of using our proposed averaging to approximate space-varying blur, we trained another network with the same architecture as in Fig.~\ref{fig:cnn}. However, we used blurry-sharp image pairs, where the blurry image is obtained synthetically via a shift-invariant convolutional model. As in \cite{ayaneccv}, we prepared a set of $10^5$ different blurs. During training, we randomly pick one of these motion blurs and convolve it with a sharp image (from a mixture of $50K$ sharp frames from our WILD dataset and $100K$ cityscapes images\footnote{www.cityscapes-dataset.com}) to generate blurred data. We refer to this trained network as the \emph{DeblurNet}$^\text{SI}$, where \emph{SI} stands for shift-invariant blur. A second network is instead trained only on the blurry-sharp image pairs from our WILD dataset (a total of $50K$ image pairs obtained from the selection and averaging process on the GoPro Hero5 Black videos). This network is called \emph{DeblurNet}$^\text{WILD}$, where \emph{WILD} stands for the data from the WILD dataset. As will be seen later in the experiments, the \emph{DeblurNet}$^\text{WILD}$ network outperforms the \emph{DeblurNet}$^\text{SI}$ network despite the smaller training set and the fact that the same sharp frames from the WILD dataset have been used. Therefore, due to space limitations, often we will show only results of the \emph{DeblurNet}$^\text{WILD}$ network in the comparisons with other methods.

\begin{figure*}[t]
  \centering
  \begin{minipage}[b]{.49\textwidth}
        \includegraphics[width= 1 \textwidth]{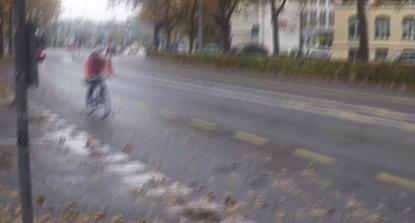}
   \end{minipage}
   \begin{minipage}[b]{.49\textwidth}
        \includegraphics[width= 1 \textwidth]{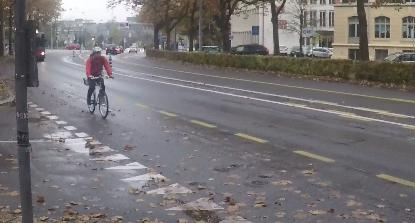}
   \end{minipage}\\\vspace{-.12cm}
(a)\hspace{.45\textwidth}(b)\\   
   \begin{minipage}[b]{.49\textwidth}
        \includegraphics[width= 1 \textwidth]{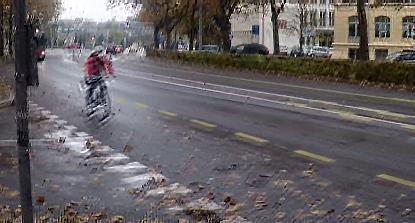}
   \end{minipage} 
   \begin{minipage}[b]{.49\textwidth}
        \includegraphics[width= 1 \textwidth]{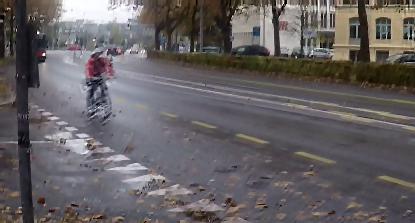}
   \end{minipage}\\\vspace{-.12cm}
(c)\hspace{.45\textwidth}(d)\\
   \begin{minipage}[b]{.49\textwidth}
        \includegraphics[width= 1 \textwidth]{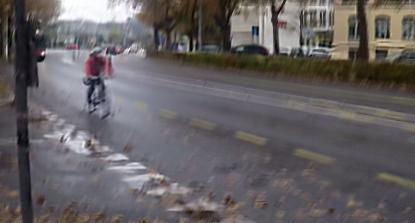}
   \end{minipage}    
   \begin{minipage}[b]{.49\textwidth}
        \includegraphics[width= 1 \textwidth]{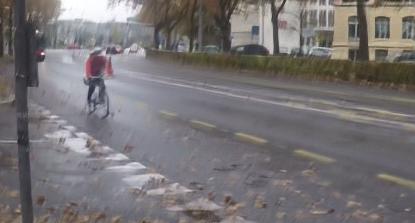}
   \end{minipage}\\\vspace{-.12cm}
   (e)\hspace{.45\textwidth}(f) 
  \label{fig:exp1}
  \caption{An example from the WILD test set. (a) blurry image, (b) sharp image (ground truth), (c) Xu and Jia \cite{Xu2010}, (d) Xu \etal \cite{xu2013unnatural}, (e) Sun \etal \cite{sun2015learning}, (f) \emph{DeblurNet}$^\text{WILD}$.\label{fig:test2}}
\end{figure*}
\begin{figure}
  \centering
  \begin{minipage}[b]{.24\textwidth}
        \includegraphics[width= 1 \textwidth]{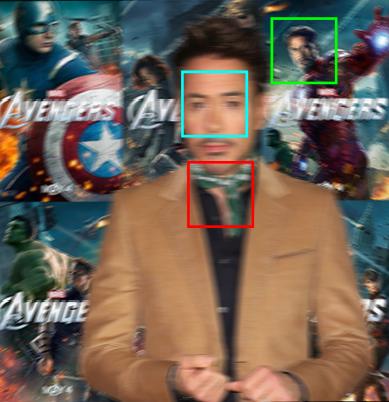}
   \end{minipage}
   \begin{minipage}[b]{.24\textwidth}
        \includegraphics[width= 1 \textwidth]{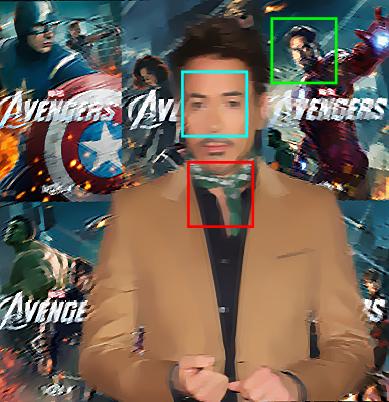}
   \end{minipage}    
   \begin{minipage}[b]{.24\textwidth}
        \includegraphics[width= 1 \textwidth]{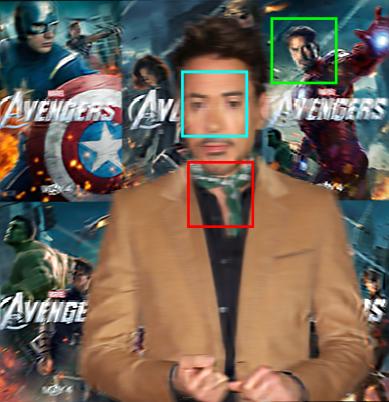}
   \end{minipage}
   \begin{minipage}[b]{.24\textwidth}
        \includegraphics[width= 1 \textwidth]{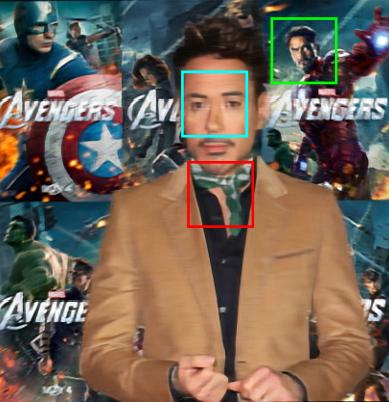}
   \end{minipage}
   
   \begin{minipage}[b]{.24\textwidth}
        \includegraphics[width=.315\textwidth]{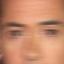}
		\includegraphics[width=.315\textwidth]{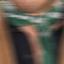}
		\includegraphics[width=.315\textwidth]{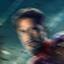}
   \end{minipage}
   \begin{minipage}[b]{.24\textwidth}
        \includegraphics[width=.315\textwidth]{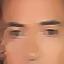}
		\includegraphics[width=.315\textwidth]{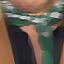}
		\includegraphics[width=.315\textwidth]{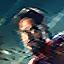}
   \end{minipage}
      \begin{minipage}[b]{.24\textwidth}
        \includegraphics[width=.315\textwidth]{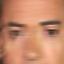}
		\includegraphics[width=.315\textwidth]{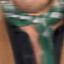}
		\includegraphics[width=.315\textwidth]{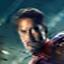}
   \end{minipage}
   \begin{minipage}[b]{.24\textwidth}
        \includegraphics[width=.315\textwidth]{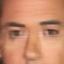}
		\includegraphics[width=.315\textwidth]{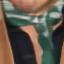}
		\includegraphics[width=.315\textwidth]{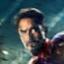}
   \end{minipage}\\\vspace{-.1cm}
  (a)\hspace{.22\textwidth}(b)\hspace{.22\textwidth}(c)\hspace{.22\textwidth}(d)\\

  \setcounter{subfigure}{0}
  \begin{minipage}[b]{.24\textwidth}
        \includegraphics[width= 1 \textwidth]{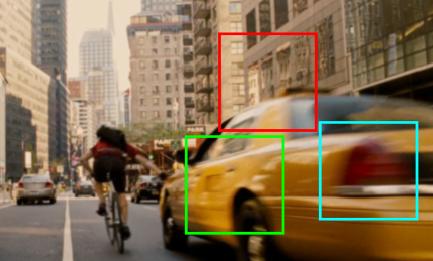}
   \end{minipage}
   \begin{minipage}[b]{.24\textwidth}
        \includegraphics[width= 1 \textwidth]{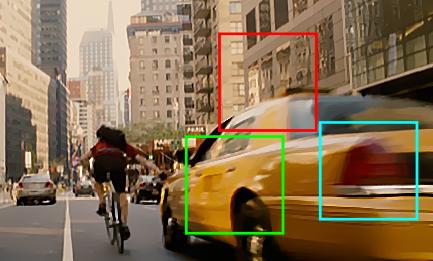}
   \end{minipage}    
   \begin{minipage}[b]{.24\textwidth}
        \includegraphics[width= 1 \textwidth]{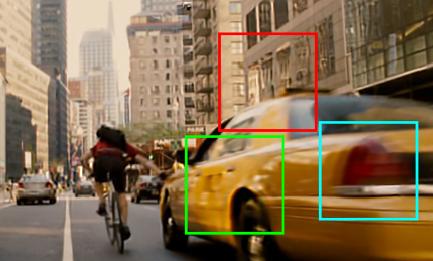}
   \end{minipage}
   \begin{minipage}[b]{.24\textwidth}
        \includegraphics[width= 1 \textwidth]{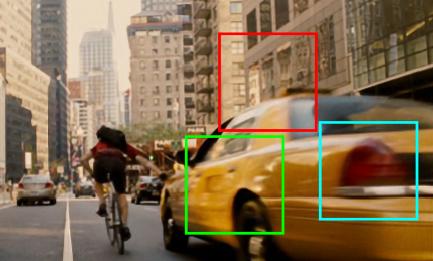}
   \end{minipage}
   
   \begin{minipage}[b]{.24\textwidth}
        \includegraphics[width=.315\textwidth]{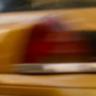}
		\includegraphics[width=.315\textwidth]{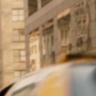}
		\includegraphics[width=.315\textwidth]{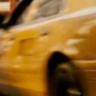}
   \end{minipage}
   \begin{minipage}[b]{.24\textwidth}
        \includegraphics[width=.315\textwidth]{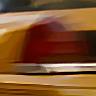}
		\includegraphics[width=.315\textwidth]{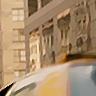}
		\includegraphics[width=.315\textwidth]{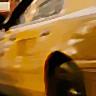}
   \end{minipage}
      \begin{minipage}[b]{.24\textwidth}
        \includegraphics[width=.315\textwidth]{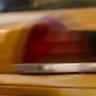}
		\includegraphics[width=.315\textwidth]{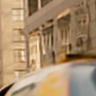}
		\includegraphics[width=.315\textwidth]{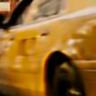}
   \end{minipage}
   \begin{minipage}[b]{.24\textwidth}
        \includegraphics[width=.315\textwidth]{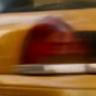}
		\includegraphics[width=.315\textwidth]{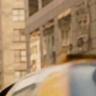}
		\includegraphics[width=.315\textwidth]{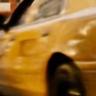}
   \end{minipage}\\\vspace{-.1cm}
  (e)\hspace{.22\textwidth}(f)\hspace{.22\textwidth}(g)\hspace{.22\textwidth}(h)
\caption{Test set from \cite{Hyun2014}. (a,e) Blurry image; (b,f)  Kim and Lee \cite{Hyun2014}; (c,g) Sun \etal \cite{sun2015learning}; (d,h) \emph{DeblurNet}$^\text{WILD}$.
\label{fig:ponce2}}
\end{figure}
\begin{figure}
  \centering
   \setcounter{subfigure}{0}
   \begin{minipage}[b]{.32\textwidth}
        \includegraphics[width= 1 \textwidth]{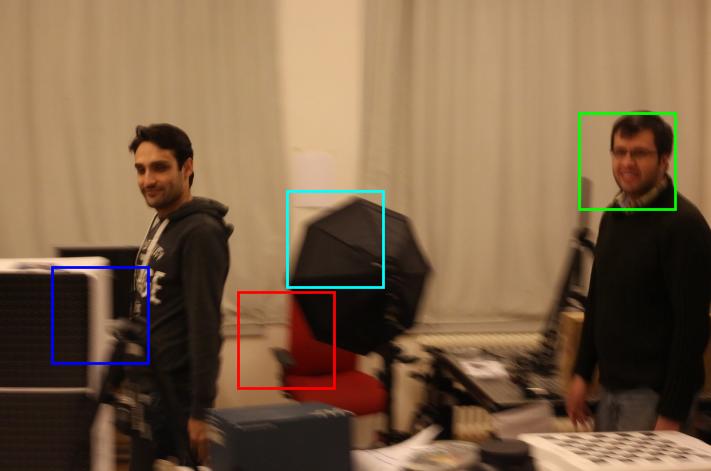}
   \end{minipage}
   \begin{minipage}[b]{.32\textwidth}
        \includegraphics[width= 1 \textwidth]{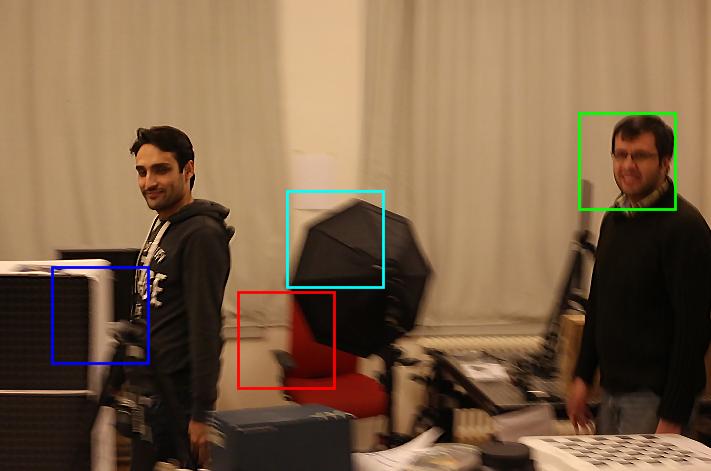}
   \end{minipage}    
   \begin{minipage}[b]{.32\textwidth}
        \includegraphics[width= 1 \textwidth]{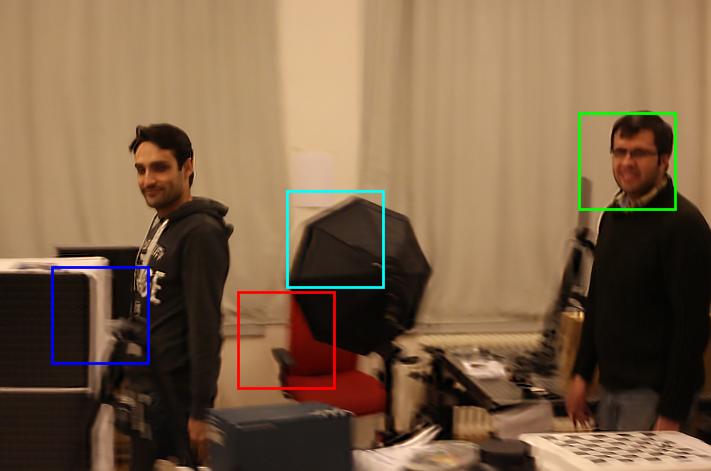}
   \end{minipage}
   
   \begin{minipage}[b]{.32\textwidth}
        \includegraphics[width=.235\textwidth]{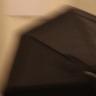}
		\includegraphics[width=.235\textwidth]{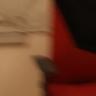}
		\includegraphics[width=.235\textwidth]{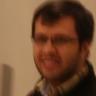}
		\includegraphics[width=.235\textwidth]{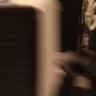}
   \end{minipage}
   \begin{minipage}[b]{.32\textwidth}
        \includegraphics[width=.235\textwidth]{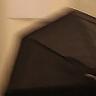}
		\includegraphics[width=.235\textwidth]{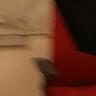}
		\includegraphics[width=.235\textwidth]{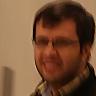}
		\includegraphics[width=.235\textwidth]{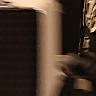}
   \end{minipage}
   \begin{minipage}[b]{.32\textwidth}
        \includegraphics[width=.235\textwidth]{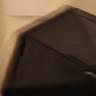}
		\includegraphics[width=.235\textwidth]{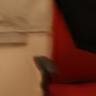}
		\includegraphics[width=.235\textwidth]{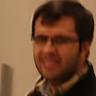}
		\includegraphics[width=.235\textwidth]{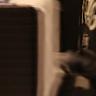}
   \end{minipage}
\\\vspace{-.1cm}
  (a)\hspace{.3\textwidth}(b)\hspace{.3\textwidth}(c)\\
   \begin{minipage}[b]{.32\textwidth}
        \includegraphics[width= 1 \textwidth]{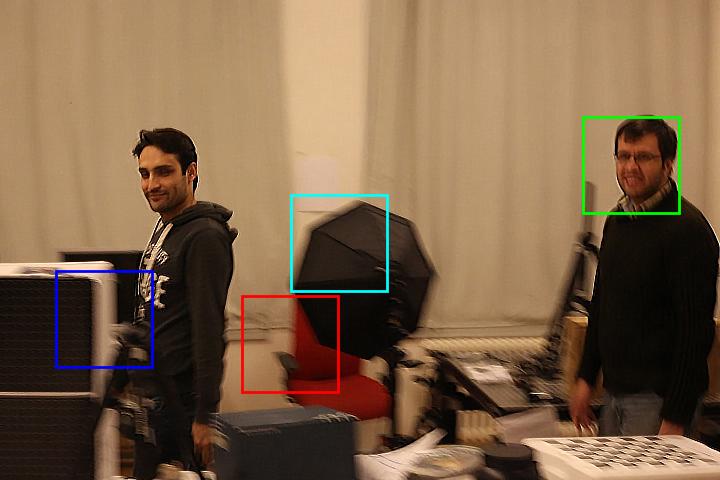}
   \end{minipage}
   \begin{minipage}[b]{.32\textwidth}
        \includegraphics[width= 1 \textwidth]{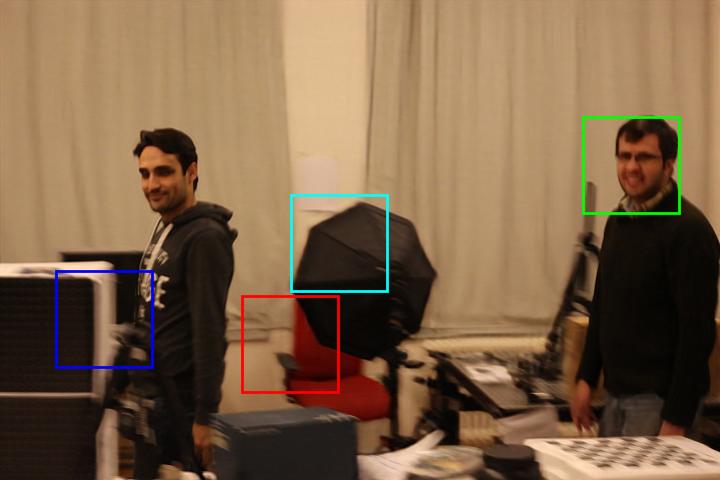}
   \end{minipage}    
   \begin{minipage}[b]{.32\textwidth}
        \includegraphics[width= 1 \textwidth]{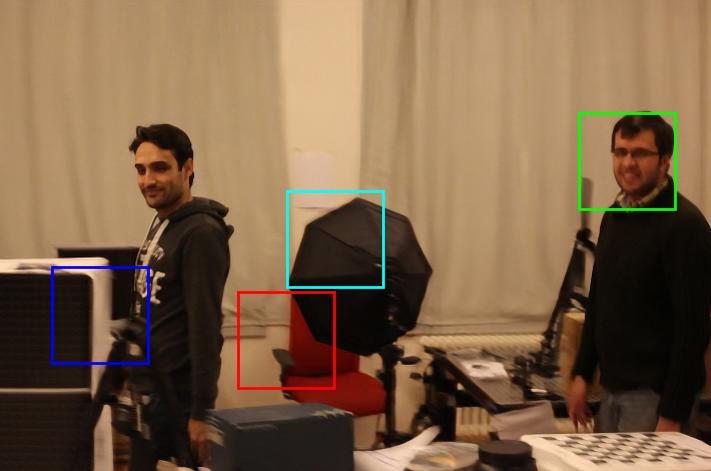}
   \end{minipage}
   
   \begin{minipage}[b]{.32\textwidth}
        \includegraphics[width=.235\textwidth]{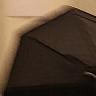}
		\includegraphics[width=.235\textwidth]{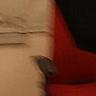}
		\includegraphics[width=.235\textwidth]{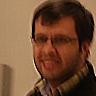}
		\includegraphics[width=.235\textwidth]{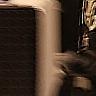}
   \end{minipage}
   \begin{minipage}[b]{.32\textwidth}
        \includegraphics[width=.235\textwidth]{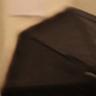}
		\includegraphics[width=.235\textwidth]{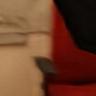}
		\includegraphics[width=.235\textwidth]{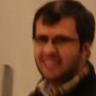}
		\includegraphics[width=.235\textwidth]{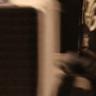}
   \end{minipage}
   \begin{minipage}[b]{.32\textwidth}
        \includegraphics[width=.235\textwidth]{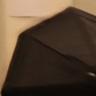}
		\includegraphics[width=.235\textwidth]{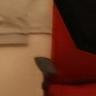}
		\includegraphics[width=.235\textwidth]{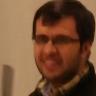}
		\includegraphics[width=.235\textwidth]{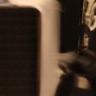}
   \end{minipage}
\\\vspace{-.1cm}
  (d)\hspace{.3\textwidth}(e)\hspace{.3\textwidth}(f) 
  \caption{Test set from the Canon camera. (a) Blurry image; (b) Xu \etal \cite{xu2013unnatural}; (c) Sun \etal \cite{sun2015learning}; (d) Xu and Jia \cite{Xu2010}; (e) \emph{DeblurNet}$^\text{SI}$; (f) \emph{DeblurNet}$^\text{WILD}$.\label{fig:canon}}
\end{figure}

\noindent\textbf{WILD test set evaluation.}
The videos in the test set were captured at locations different from those where training data was captured. Also, incidentally, the weather conditions during the capture of the test set were significantly different from those of the training set. We randomly chose $15$ images from the test-set and compared the performance of our method against the methods in \cite{Xu2010}, \cite{sun2015learning}, the space-varying implementation of the method in \cite{xu2013unnatural}, and \emph{DeblurNet}$^\text{WILD}$ trained network. An example image is shown in Fig.~\ref{fig:test2}. 
As can be observed, blur variation due to either object motion or depth changes is the major cause of artifacts. Our \emph{DeblurNet}$^\text{WILD}$ network, however, produces artifact-free sharp images.
While the example in Fig.~\ref{fig:test2} gives only a qualitative evaluation, in Table~\ref{tbl:psnr} we report quantitative results. 

\begin{wraptable}{r}{7.5cm}
\begin{tabular}{ @{\hspace{0em}} c @{\hspace{.7em}} c @{\hspace{.7em}}  c @{\hspace{.7em}} c @{\hspace{.7em}} c @{\hspace{0em}} }
\hline
\cite{sun2015learning} & \cite{Xu2010} & \cite{xu2013unnatural} & \emph{DeblurNet}$^\text{SI}$ & \emph{DeblurNet}$^\text{WILD}$ \\
\hline 
$25.48$ & $23.61$ & $22.50$ & $25.8$ & $\mathbf{28.1}$ \\
\hline \\
\end{tabular}\vspace{-.5cm}
\caption{Average PSNR on our WILD test set.\label{tbl:psnr}} 
\end{wraptable} 
We measure the performance of all the above methods in terms of Peak Signal-to-Noise Ratio (PSNR) by using the reference sharp image as in standard image deblurring performance evaluations. We can see that the performance of the \emph{DeblurNet}$^\text{WILD}$ is better than that of the \emph{DeblurNet}$^\text{SI}$. This is not surprising because the shift-invariant training set does not capture factors such as reflections/specularities, the space-varying blur, occlusions and coming into view of objects. Notice that the PSNR values are not comparable to those seen in shift-invariant deconvolution algorithms. 

\begin{figure*}[t]
  \centering
  \begin{minipage}[b]{.325\textwidth}
        \includegraphics[width= 1 \textwidth]{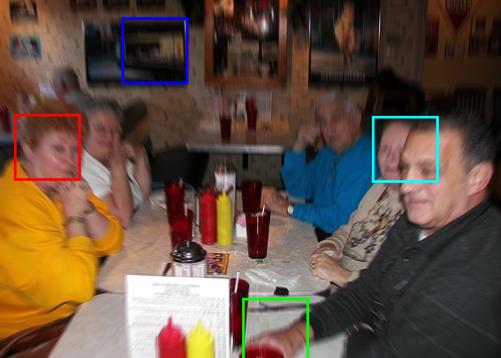}
   \end{minipage}
   \begin{minipage}[b]{.325\textwidth}
        \includegraphics[width= 1 \textwidth]{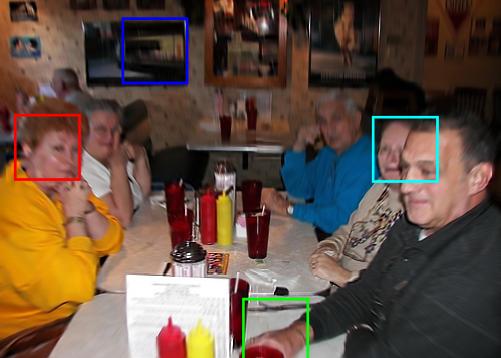}
   \end{minipage}    
   \begin{minipage}[b]{.325\textwidth}
        \includegraphics[width= 1 \textwidth]{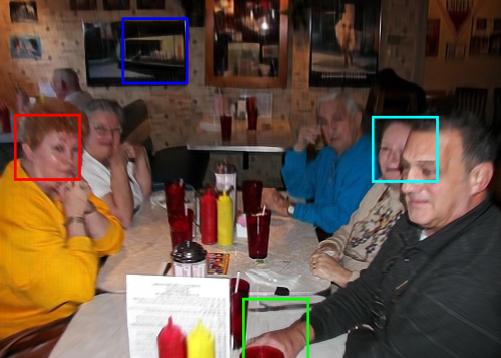}
   \end{minipage}
   \begin{minipage}[b]{.325\textwidth}
        \includegraphics[width=.235\textwidth]{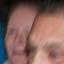}\hfill\includegraphics[width=.235\textwidth]{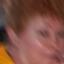}\hfill\includegraphics[width=.235\textwidth]{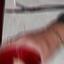}\hfill\includegraphics[width=.235\textwidth]{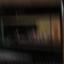}
   \end{minipage}
   \begin{minipage}[b]{.325\textwidth}
        \includegraphics[width=.235\textwidth]{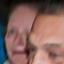}\hfill\includegraphics[width=.235\textwidth]{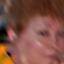}\hfill\includegraphics[width=.235\textwidth]{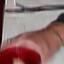}\hfill\includegraphics[width=.235\textwidth]{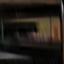}
   \end{minipage}
   \begin{minipage}[b]{.325\textwidth}
        \includegraphics[width=.235\textwidth]{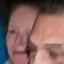}\hfill\includegraphics[width=.235\textwidth]{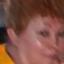}\hfill\includegraphics[width=.235\textwidth]{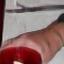}\hfill\includegraphics[width=.235\textwidth]{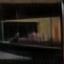}
   \end{minipage}\\\vspace{-.1cm}
  (a)\hspace{.3\textwidth}(b)\hspace{.3\textwidth}(c) 
  \caption{Test dataset from \cite{sun2015learning}. (a) Blurry image, (b) Sun \etal\cite{sun2015learning}, (c)  \emph{DeblurNet}$^\text{WILD}$.\label{fig:ponce1}}
\end{figure*}

\noindent\textbf{Qualitative evaluation.}
On other available \emph{dynamic scene blur} datasets the ground truth is not available. Therefore, we can only evaluate our proposed network qualitatively. We consider $2$ available datasets and images obtained from a Canon EOS 5D Mark II camera. While Figs.~\ref{fig:ponce2} and \ref{fig:ponce1} show data from \cite{sun2015learning} and \cite{Hyun2014} respectively, Fig.~\ref{fig:canon} shows images from the Canon camera. 
In Fig.~\ref{fig:canon}, we compare the methods of \cite{Xu2010}, \cite{sun2015learning} and \cite{xu2013unnatural} to both our \emph{DeblurNet}$^\text{SI}$ and \emph{DeblurNet}$^\text{WILD}$ networks. In all datasets, we observe that our method is able to return sharper images with fine details. Furthermore, we observe that in Fig.~\ref{fig:canon} the \emph{DeblurNet}$^\text{WILD}$ network produces better results than the \emph{DeblurNet}$^\text{SI}$ network, which confirms once more our expectations.

\begin{figure}[t]
      \includegraphics[height= .185 \textwidth]{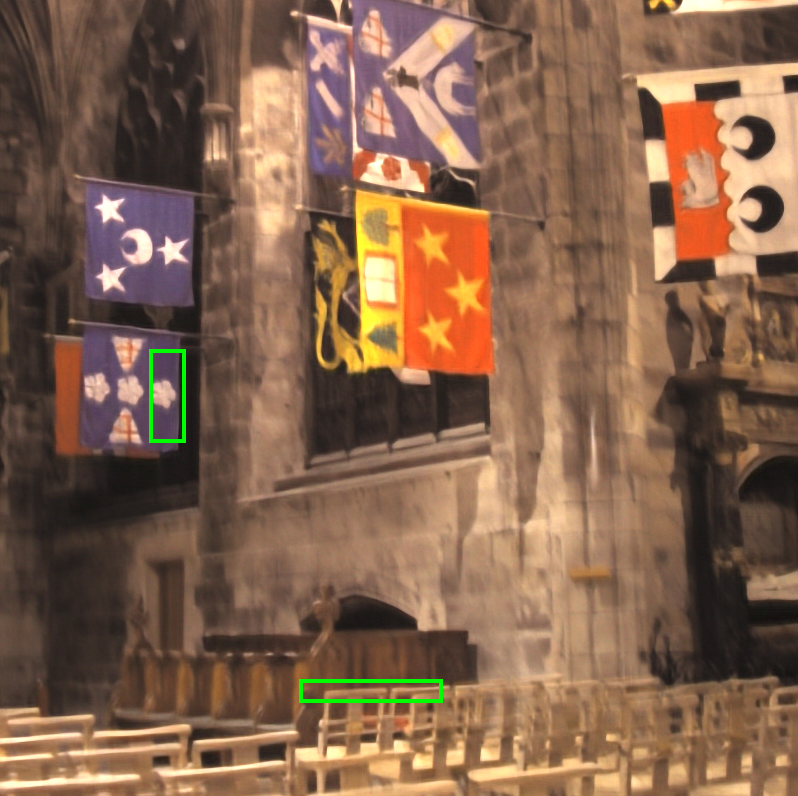}
      \includegraphics[height= .185 \textwidth]{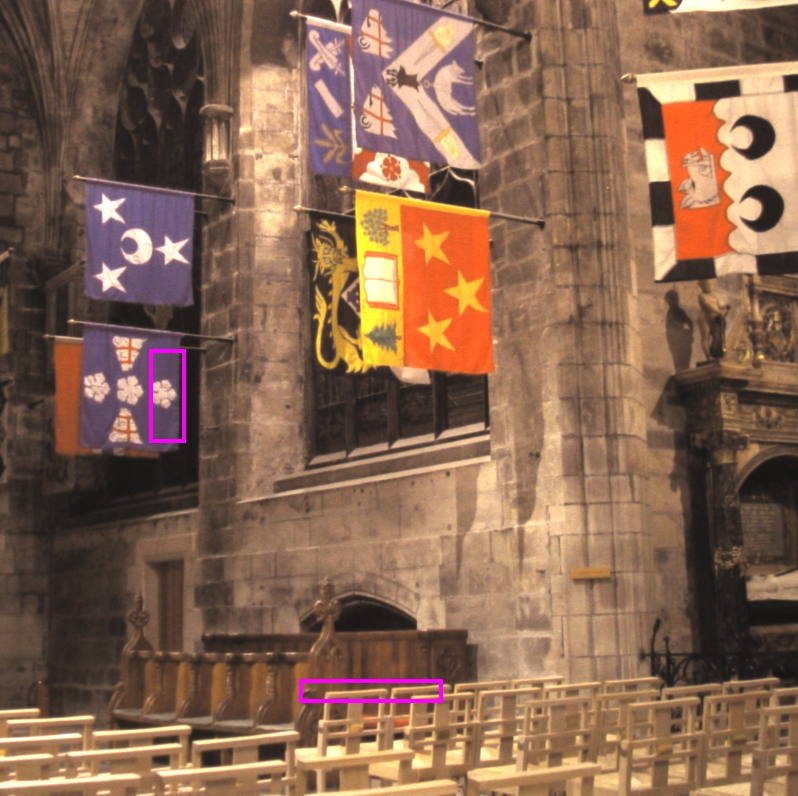}
	  \includegraphics[height= .185 \textwidth]{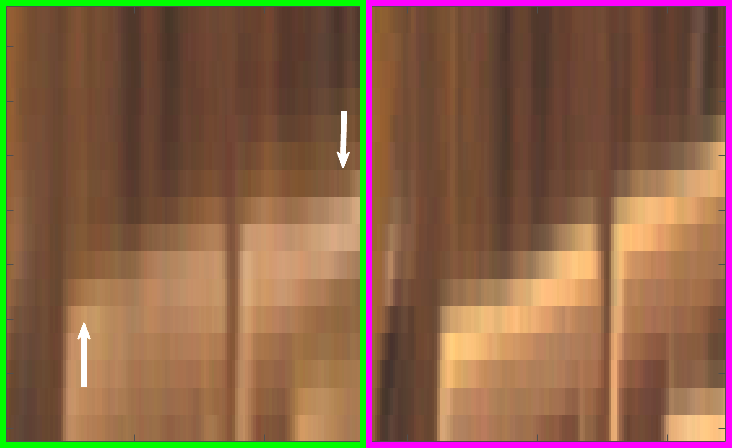}
      \includegraphics[height= .185 \textwidth]{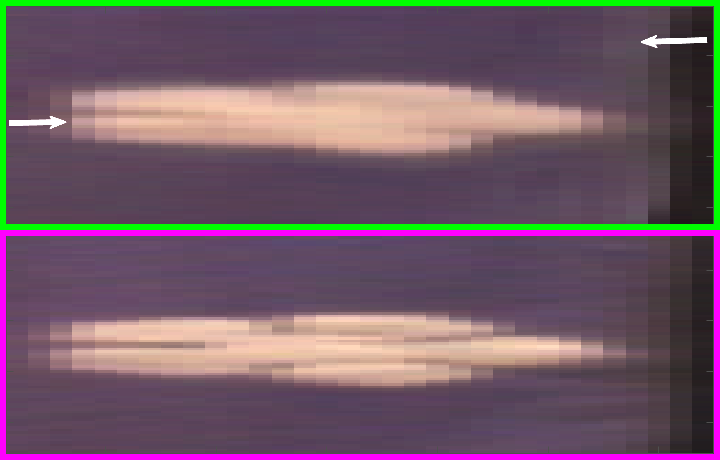}\\\vspace{-.1cm}
  (a)\hspace{.17\textwidth}(b)\hspace{.22\textwidth}(c)\hspace{.27\textwidth}(d)\hspace{.06\textwidth} 
   \caption{Kohler dataset \cite{Kohler2012} (image 1, blur 4). (a) our result. (b) ground truth. (c,d) Zoomed-in patches. Local ambiguous shifts are marked with white arrows.}\label{fig:shift}
\end{figure}

\noindent\textbf{Shift-invariant blur evaluation.}
We provide a brief analysis on the differences between dynamic scene deblurring and shift-invariant motion deblurring. We use an example from the standard dataset of \cite{Kohler2012}, where blur is due to camera shake (see Fig.~\ref{fig:shift}). 
In the case of a shift-invariant blur, there are infinite $\{$blur, sharp image$\}$ pairs that yield the same blurry image when convolved. More precisely, an unknown 2D translation (shift) in a sharp image $f$ can be compensated by an opposite 2D translation in the blur kernel $k$, that is, $\forall\Delta$, 
$g(x) = \int f(y+\Delta)k(x-y-\Delta) dy.$
Because of such ambiguity, current evaluations compute the PSNR for all possible 2D shifts of $f$ and pick the highest PSNR. The analogous search is done for camera shake \cite{Kohler2012}. However, with a dynamic scene we have ambiguous shifts at every pixel (see Fig.~\ref{fig:shift}) and such search is unfeasible (the image deformation is undefined).
Therefore, all methods for dynamic scene blur would be at a disadvantage with the current shift-invariant blur evaluation methods, although their results might look qualitatively good.
\begin{figure}[t]
\includegraphics[trim = {5cm 1cm 2cm 2cm}, width= .5 \textwidth]{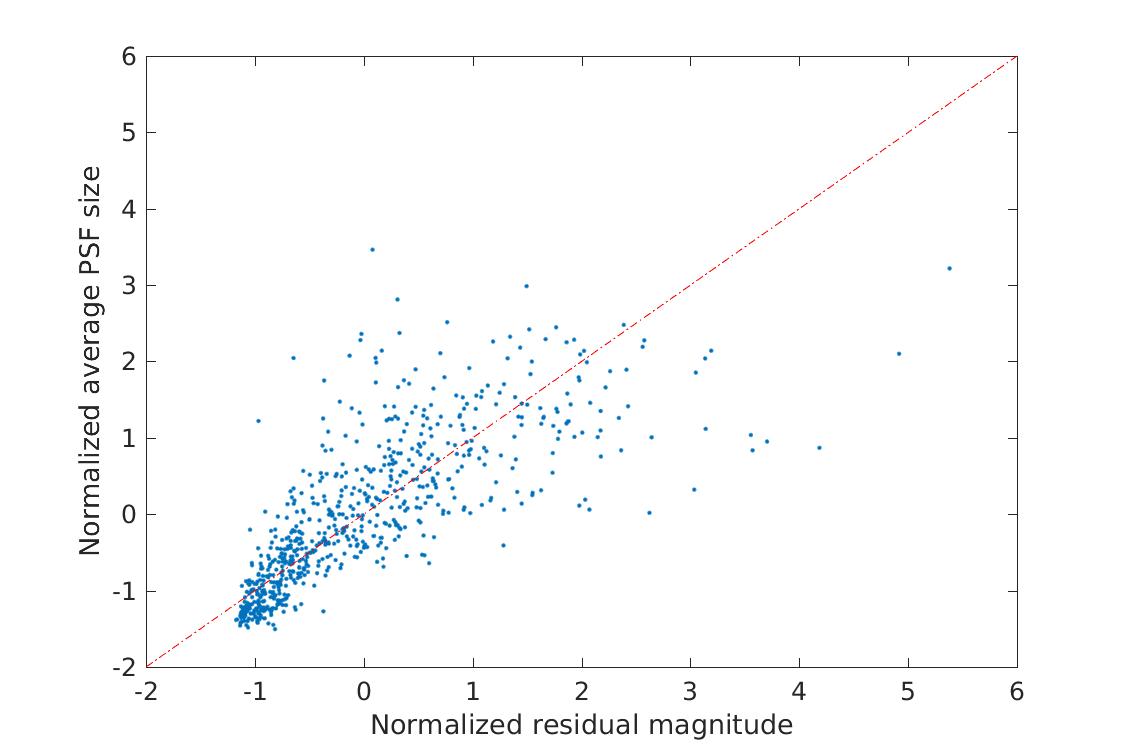}
\caption{Normalized average blur size versus normalized residual magnitude plot. Notice the high level of correlation between the blur size and the residual magnitude. \label{fig:plot}}
\end{figure}
\begin{figure}
  \centering
  \begin{minipage}[b]{.245\textwidth}
        \includegraphics[width= 1 \textwidth]{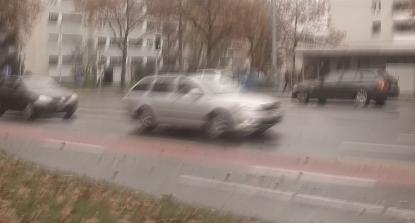}
   \end{minipage}
   \begin{minipage}[b]{.245\textwidth}
        \includegraphics[width= 1 \textwidth]{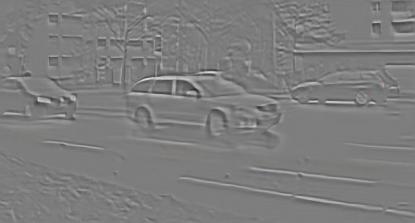}
   \end{minipage}    
   \begin{minipage}[b]{.245\textwidth}
        \includegraphics[width= 1 \textwidth]{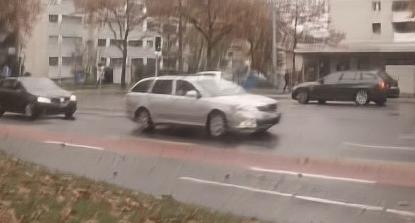}
   \end{minipage}
   \begin{minipage}[b]{.245\textwidth}
        \includegraphics[width= 1 \textwidth]{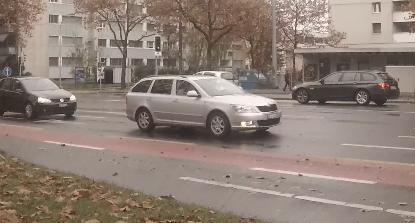}
   \end{minipage}
   \begin{minipage}[b]{.245\textwidth}
        \includegraphics[width= 1 \textwidth]{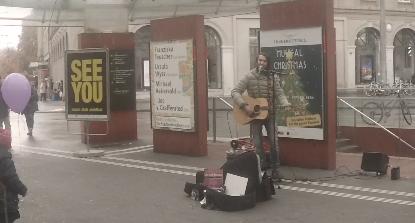}
   \end{minipage}
   \begin{minipage}[b]{.245\textwidth}
        \includegraphics[width= 1 \textwidth]{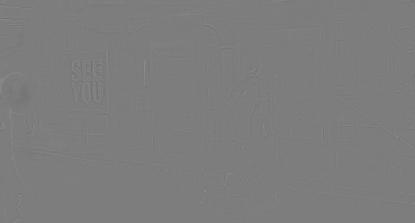}
   \end{minipage}    
   \begin{minipage}[b]{.245\textwidth}
        \includegraphics[width= 1 \textwidth]{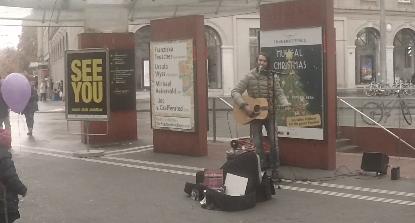}
   \end{minipage}
   \begin{minipage}[b]{.245\textwidth}
        \includegraphics[width= 1 \textwidth]{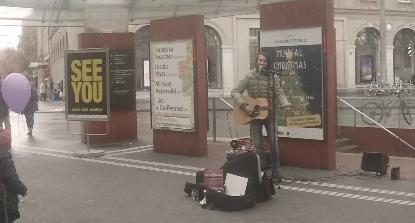}
   \end{minipage}
  \caption{The images with highest (first row) and lowest (second row) residual norm in the output layer. The image in the first column is the input, the second column shows the estimated residual (the network output), the third column is the deblurred image (first column + second column), and finally the forth column is the ground truth.
\label{fig:analysis}}
\end{figure}

\noindent\textbf{Analysis.}
Our network generates a residual image that when added to the blurry input yields the sharp image. 
Therefore, we expect the magnitude of the residual to be large for very blurry images, as more changes will be required. 
To validate this hypothesis we perform both quantitative and qualitative experiments.
We take $700$ images from another WILD test set (different from the $15$ images used in the previous quantitative evaluation), provide them as input to the \emph{DeblurNet}$^\text{WILD}$ network, and calculate the $L^1$ norm of the network residuals (the output of the last layer of $N_3$). In Fig.~\ref{fig:analysis} we show two images, one with the highest and one with the lowest $L^1$ norm. We see that the residuals with the highest norms correspond to highly blurred images, and vice versa for the low norm residuals.
We also show quantitatively that there is a clear correlation between the amount of blur and the residual $L^1$ norm. As mentioned earlier on, our WILD dataset also computes an estimate of the blurs by integrating the optical flow. We use this blur estimate to calculate the average blur size across the blurry image. This gives us an approximation of the overall amount of blur in an image. In Fig.~\ref{fig:plot} we show the plot of the $L^1$ norm of the residual versus the average estimated blur size for all $700$ images. The residual magnitudes and blur sizes are normalized so that mean and standard deviation are $0$ and $1$ respectively. 

\section{Conclusions}
We proposed DeblurNet, a novel CNN architecture that regresses a sharp image given a blurred one. DeblurNet is able to restore blurry images under challenging conditions, such as occlusions, motion parallax and camera rotations. The network consists of a chain of $3$ subgraphs, which implement a multiscale strategy to break down the complexity of the deblurring task. Moreover, each subgraph outputs only a residual image that yields the sharp image when added to the input image. This allows the subgraph to focus on small details as confirmed experimentally. An important part of our solution is the design of a sufficiently realistic dataset. We find that simple frame averaging combined with a very high frame-rate video camera produces reasonable blurred-sharp image pairs for the training of our DeblurNet network. Indeed, both quantitative and qualitative results show state of the art performance when compared to prior dynamic scene deblurring work. We observe that our network does not generate artifacts, but may leave extreme blurs untouched. 

{\smallskip
\noindent\textbf{Acknowledgements.}
Paolo Favaro acknowledges support from the Swiss National Science Foundation on project 200021\_153324.
}
  
	\bibliographystyle{splncs03}
	\bibliography{papers}

\end{document}